\documentclass[10pt,twocolumn,letterpaper]{article}

\usepackage{iccv}
\usepackage{times}
\usepackage{epsfig}
\usepackage{graphicx}
\usepackage{amsmath}
\usepackage{amssymb}
\usepackage{multirow}

\usepackage[hyphens]{url}
\usepackage[hyphenbreaks]{breakurl}

\usepackage[breaklinks=true,bookmarks=false]{hyperref}

\iccvfinalcopy 

\def\iccvPaperID{****} 

\ificcvfinal\fi

\begin{document}

\title{2nd Place Solution to Google Landmark Retrieval 2021}

\author{Yuqi Zhang, Xianzhe Xu, Weihua Chen, Yaohua Wang, Fangyi Zhang, Fan Wang, Hao Li\\
Alibaba Group\\
{\tt\small \{gongyou.zyq, xianzhe.xxz, kugang.cwh, xiachen.wyh, zhiyuan.zfy,} \\
{\tt\small fan.w, lihao.lh\}@alibaba-inc.com}
}

\maketitle
\ificcvfinal\thispagestyle{empty}\fi

\begin{abstract}
This paper presents the 2nd place solution to the Google Landmark Retrieval 2021 Competition on Kaggle. The solution is based on a baseline with training tricks from person re-identification, a continent-aware sampling strategy is presented to select training images according to their country tags and a Landmark-Country aware reranking is proposed for the retrieval task. With these contributions, we achieve 0.52995 mAP@100 on private leaderboard. Code available at \url{ https://github.com/WesleyZhang1991/Google_Landmark_Retrieval_2021_2nd_Place_Solution}
\end{abstract}

\section{Introduction}

Image retrieval is a very important computer vision task which aims at finding images similar to the query image. It is different from instance-level retrieval. Image retrieval aims to retrieve objects holding the same appearance with the query, even they are not the same instance. This makes the task more easier compared to instance-level retrieval. 
On the other hand, it is different from the fine-grained level image retrieval. The fine-grained level image retrieval pays more attention on the local attentions to discover more details due to its small intra-category variance, such as person re-identification.

Landmark retrieval is an instance-level retrieval task, which aims to search the same landmark from a large candidate set. In this paper, we will introduce our techniques used in the fourth landmark retrieval competition, Google Landmark Retrieval 2021~\cite{ILR2021} held on Kaggle. Some of them are inspired from the state of the art algorithms in person re-identification.

Besides, we also involve many techniques that are commonly used in previous competitions~\cite{jeon20201st, mei20203rd, ozaki2019large, henkel2020supporting, dai2020}, including model structures~\cite{tan2019efficientnet}, training strategies, loss functions~\cite{deng2019arcface, chen2017multi, chen2017beyond, zhang2019cross}. These techniques have been well explored and introduced in previous competitions, so we only introduce our methods, denoted as new contributions listed below.
\begin{itemize}
\item We involve bags-of-tricks from person re-identification and conduct careful experiments on these tricks.

\item We propose a continent-aware sampler to balance the distribution of training images based on their continent tags.

\item We design a Landmark-Country aware reranking algorithm and integrate it with the K-reciprocal reranking method.
\end{itemize}

\section{Method}

\subsection{Baseline network}
We select several large CNN networks including SE-ResNet-101~\cite{hu2018squeeze}, ResNeXt-101~\cite{xie2017aggregated}, ResNeSt101 and ResNeSt269~\cite{zhang2020resnest} as backbones. IBN extension~\cite{pan2018two} is used for SE-ResNet and ResNeXt-101. The input size is selected as 384 for pretraining and 512 for the last fine-tuning. The last stride of the CNN network is set to 1.
We use generalized mean-pooling (GeM) for pooling method with p=3.0. Arcface loss~\cite{deng2019arcface} with $scale=30$ and $margin=0.3$ is used. We use weight decay=0.0005. Training details with gradually enlarging input size and data scale can be found in Section~\ref{implementation_detail}.

Some tricks from person re-identification has been explored. Random Erasing~\cite{zhong2020random} randomly erases out image patches and has shown great success in many fields. Label smoothing~\cite{muller2019does} by using soft targets that are a weighted average of the hard targets can often be useful in many computer vision tasks.

\subsection{Sampling Strategy}
On one hand, in person re-identification or face recognition, id-uniform is widely used as a data sampling strategy. For a batch, we randomly select P Ids and then K images for each Id. Thus we have P*K images as a batch. Each id is treated fair for this setting. On the other hand, softmax sampling has been widely used in previous competitions. The softmax sampling just shuffle all dataset once at the beginning of the epoch and then samples iterative through the data. Head data which appears more will be put more attention with this setting.
We have tried these sampling strategies and find neither of them are good enough for our task.

As the provided landmark dataset pays more concentration on Asia landmarks~\cite{kim2021towards}, we manage to design a sampling strategy based on their continent labels. 
First, we use the country-and-continent-codes-list~\cite{link1} to find how many countries each continent has. Then we search and list all the landmarks in each country from~\cite{link2}. Based on this processing, we can get the country tag and continent tag for every landmark.

We setup a continent sampling prob by {'Asia': 0.5, 'Europe': 0.2, 'Africa': 0.15, 'North America': 0.1, 'South America': 0.02, 'Antarctica': 0.01, 'Oceania': 0.01, 'OTHER': 0.01}. For an epoch of images, we sample continent images by the corresponding ratios. Also, we learn from previous paper~\cite{jeon20201st} which set 0.66 probability for clean data and 0.33 probability for noisy data.

\subsection{Reranking}
Reranking is very essential to the final performance. Besides K-reciprocal reranking~\cite{zhong2017re}, a Landmark-Country aware reranking algorithm is specially designed for our task.

We observe that there are many images which contain the same landmark and can't be easily retrieved by visual features due to great variations caused by views and illumination. Considering this, a Landmark-Country aware reranking is proposed by taking fully use of the training set. Specifically, as each image in training set has its landmark tag and its country tag, we first assign query and index images in the testing set with a training tag, and then retrieval the query from index images by the assigned tags. 

For query image tagging, We give a list of potential landmark tags and country tags to every query image according to its top K similar images in training set, and each landmark tag and country tag is scored by the accumulation of similarity in top k, as shown in Equation~\eqref{equa1}. The $j$th tag appears in topk is noted as $score_j$. For each index image, the landmark tag and country tag are given by its most similar image in training set. 
\begin{equation}
    score_j = \sum_{i}^{k} sim_{i==j} \label{equa1}
\end{equation}

Then we reranking the query's retrieval results using Equation ~\eqref{equa2}. 
\begin{equation}
    sim_{final} = sim + \alpha \times L_{score} + \beta \times C_{score} \label{equa2}
\end{equation}
where $\alpha$ and $\beta$ is the weight of the landmark tag score and the country tag score respectively.

With the Landmark-Country aware reranking, images have the same landmark tag or country tag with the potential tag of query are advanced in the retrieval sorting. 

\section{Experiments}
\subsection{Training set construction}
The official GLDv2 dataset~\cite{weyand2020google} has provided a clean and a full version. The clean version include clean data, while the full version contain both clean and noise data. As pointed out by previous works~\cite{dai2020}, many noisy images in the full set with the labels of clean set have been filtered out during the cleaning stage. Expanding these noisy data into clean set results in `c2x`. Although the existing of noise, these noisy images may contain very valuable information, e.g., the indoor or outdoor of a building. What's more, we also include the index set from GLDv2, which shares many common ids with the full set. We list the dataset in Table~\ref{tab:train_data} for a better understanding.

\begin{table}[t]
\begin{center}
\begin{tabular}{|c|cc|}
\hline
Trainset 			& \# Samples 	 & \# Labels	 \\
\hline
clean               & 1,580,470      & 81,313    \\
c2x                 & 3,223,078     & 81,313   \\
full            & 4,132,914      & 203,094   \\
all                 & 4,825,830      & 203,094   \\

\hline
\end{tabular}
\end{center}
\caption{Training set construction.\label{tab:train_data}}
\end{table}

\subsection{Validation Set}
We use the 1129 queries in GLDv2 test set together with the 76,176 index set from the competition. We put all GT images of each query into the index set and expand it to 78,959 images. In this case, any query could find all its GTs in the index set.

\subsection{Implementation details}
\label{implementation_detail}
Motivated by previous solutions, we first train the models on `clean` subset with an input size of 384(For ResNeSt269, the size is 448). The initial learning rate is 0.01 and we train for 10 epochs with the first epoch as warmup. Then we keep the same category number but with more data as `c2x`. The initial learning rate is 0.001 and we train for 6 epochs. Then we expand dataset to `full` and train for another 2 epochs with the initial learning rate as 0.0001 and input size as 512. At last, we use `all` data from GLDv2 and train for another 2 epochs with the initial learning rate as 0.0001 and input size as 512.

\subsection{Bag-of-tricks}
Many tricks from person re-identification have been tested and we list some important ones in Table~\ref{tab:tricks}. Since the dataset is very large, we use R50 backbone with an input size of $256\times 256$. We list the validation accuracy as well as public/private scores. Ransom erasing proves to be effective while label smoothing fails.

\begin{table}[t]
\begin{center}
\begin{tabular}{|c|ccc|}
\hline
Setting 			& Validation	         & Public      & Private \\
\hline
baseline 			& 32.60           	     & 28.44       & 30.59   \\
+RE 				& 32.78	                 & 29.29       & 30.60 \\
+label smooth 		& 32.55           	     & 28.21       & 29.79   \\
\hline
\end{tabular}
\end{center}
\caption{Effect of different tricks on landmark retrieval.\label{tab:tricks}}
\end{table}

\subsection{Sampling strategy}
Table~\ref{tab:sampling} lists results for different sampling strategy. The widely used id-uniform strategy fails for landmark retrieval. We think the reason is due to the large amount of tail data. These tail data may be noisy and thus drifts the model training. The softmax strategy works better than id-uniform. For continent-aware strategy, it achieves the best performance.

\begin{table}[t]
\begin{center}
\begin{tabular}{|c|ccc|}
\hline
Setting 			& Validation	         & Public      & Private  \\
\hline
id-uniform 			& 31.05 			    & 24.78	       & 27.28    \\
softmax 			& 32.60                 & 28.44        & 30.59    \\
continent-aware 	& 33.07  		        & 31.37        & 32.44	  \\
\hline
\end{tabular}
\end{center}
\caption{Effect of different sampling strategy.\label{tab:sampling}}
\end{table}

\section{Conclusion}
In this paper, we import techniques and bag-of-tricks from person re-identification. For the specific task of landmark retrieval, we propose continent-aware sampling strategy and Landmark-Country aware post processing, which has proven to be very effective on the private leaderboard. The relationship between landmark retrieval and landmark recognition would be studied in the future.

{\small
\bibliographystyle{ieee_fullname}
\bibliography{egbib}
}

\end{document}